\documentclass[Journal]{IEEEtran}

\usepackage{graphicx}
\usepackage{nopageno}

\begin{document}

\title{Cognitive Computing to Optimize IT Services}

\author{\IEEEauthorblockN{Abbas Raza Ali} \\
\IEEEauthorblockA{Cognitive and Analytics Practice Leader, IBM\\
abbas.raza.ali@gmail.com}}




\maketitle
\thispagestyle{empty}

\begin{abstract}
In this paper, the challenges of maintaining a healthy IT operational environment have been addressed by proactively analyzing IT Service Desk tickets, customer satisfaction surveys, and social media data. A Cognitive solution goes beyond the traditional structured data analysis by deep analyses of both structured and unstructured text. The salient features of the proposed platform include language identification, translation, hierarchical extraction of the most frequently occurring topics, entities and their relationships, text summarization, sentiments, and knowledge extraction from the unstructured text using Natural Language Processing techniques. Moreover, the insights from unstructured text combined with structured data allow the development of various classification, segmentation, and time-series forecasting use-cases on the incident, problem, and change datasets. Further, the text and predictive insights together with raw data are used for visualization and exploration of actionable insights on a rich and interactive dashboard. However, it is hard not only to find these insights using traditional structured data analysis but it might also take a very long time to discover them, especially while dealing with a massive amount of unstructured data. By taking actions on these insights, organizations can benefit from a significant reduction of ticket volume, reduced operational costs, and increased customer satisfaction. In various experiments, on average, upto 18-25\% of yearly ticket volume has been reduced using the proposed approach. 

\end{abstract}

\begin{IEEEkeywords}
Knowledge Extraction; Optimizing IT Services; Cognitive Computing; Topic Clustering; Semantic Text Analytics; Service Desk.
\end{IEEEkeywords}

\IEEEpeerreviewmaketitle
\section{Introduction}

\IEEEPARstart{W}{ith} the increasing complexity of Information Technology's (IT) operational environments, organizations have build a need to maintain healthy IT systems while reducing operational cost, ticket volume and improving customer satisfaction at the same time. The IT operations are managed by Incident, Problem and Change (IPC) ticketing system where an issue is considered as an IT Service Desk ticket~\cite{Jan2015}. In this paper, these challenges have been addressed by utilizing structured and unstructured ticket information provided by the Service Desk (SD).

Cognitive computing offers the means to transform and scale-out IT Services Delivery by eliminating the need for repeated training of the services delivery skills. The learning ability of the system provides a key differentiator to allow the cognitive solutions to continuously improve with the on-going interactions and adapt to the changing needs. Additionally, the quality of service can be improved by avoiding errors of omission as well as a faster resolution of the problems through automated understanding and reasoning of the most valid solutions. The human element, while significantly reduced in the effort, will however continue to play a key supervisory role to provide for undisruptive user interaction. In all, cognitive computing transformation provides unsurpassed client experience and business competitiveness at operational efficiency and agility that have not yet been possible to date.

The SD receives tickets automatically generated by various device monitoring systems. Also, the SD agents manually create tickets in response to the users' requests. There are several analytical tools in the market place today that provide reports based on only the structured data attributes of historical ticket records. Some of these tools have the query capabilities to also provide the most frequently occurring keywords or specified patterns in the unstructured attributes of the tickets. These reports are then analyzed by specialized IT professionals with the objective to remedy those revealed as the most predominant topics. These tools surely provide benefits but with few limitations such as capabilities to process vast amounts of data, inability to reveal obvious and hidden issues, segregate the noise from the text, and extract users' sentiments related to key topics from Customer Satisfaction (CSat) surveys and social media data. Existing surveys in the literature estimates that 80\% of information resides in unstructured data \cite{Rosu2012}. The proposed approach is a solution that addresses all the mentioned limitations. 

A key differentiator of the proposed approach is the ability to extract meaning from fragmented sentences in the unstructured ticket description. Generally, this unstructured text is in multiple languages which is detected using language identification and normalized into a common language, English, using machine translation techniques. Conversely, due to different language skill levels of the globally sourced pool of agents, the grammar quality also varies. 

In order to develop a common baseline for the application of analytics, this solution first translates all the non-English text into English, using statistical translation models. The use of a statistical versus a grammar-based model is necessitated due to the grammar quality of the input text. The statistical translation model adds an additional level of complexity for a cognitive system. 

The system has enabled a number of clients to uncover patterns and trends for identification of the contributing causes and prescription of precautions within no time, using data from hundreds of thousands of SD tickets. Typically these types of analyses would have taken the SD support teams weeks or sometimes months. Hence, the identification of contributing causes enables the support teams to leverage log analytics for root cause analysis. Additionally, this allows the team to take a proactive stance toward identifying opportunities to optimize the IT operations and improving client satisfaction by demonstrating the following outcomes:

\begin{itemize}
\item Identification of contributing causes and actionable precautions from problem description and resolution.
\item Transparency to pervasive issues otherwise hidden behind volumes of unstructured data in tickets under different categories.
\item Identification of emerging problems, trends and automated change point detection.
\item Identification of patterns that can enable IT support teams to leverage real-time machine and data analytics to find the rapid root cause.
\item Identification of self-service and automation opportunities, and transform operations from reactive to proactive. 
\item Transparency to pervasive issues otherwise hidden behind volumes of unstructured text not captured in ticket categories.
\item Sentiment analysis of the CSat survey comments to determine potential future dissatisfiers.
\item Social Media and CSat sentiments correlation with IT and applications' issues to improve client experience.
\item Predictive Analytics for improving resolution times by re-prioritization and routing of the tickets.
\item Identification of skills and knowledge gaps derived from the analysis of first time resolution of failures, resolution time and agent performance differences.
\end{itemize}

The paper is organized as follows. Section \ref{sec:Methodology} is devoted to an overview of the cognitive computing platform components and methodology of the overall system. Section \ref{sec:Algorithms} outlines key algorithms developed for this system. Section \ref{sec:Applications} describes various applications and use-cases whereas results of various experiments are presented in Section \ref{sec:Results}. The paper is concluded in Section \ref{sec:Conclusion}.

\section{Methodology} \label{sec:Methodology}
\begin{figure}[!htbp]
\centering
\includegraphics[viewport=200 40 745 530, clip, scale = 0.5]{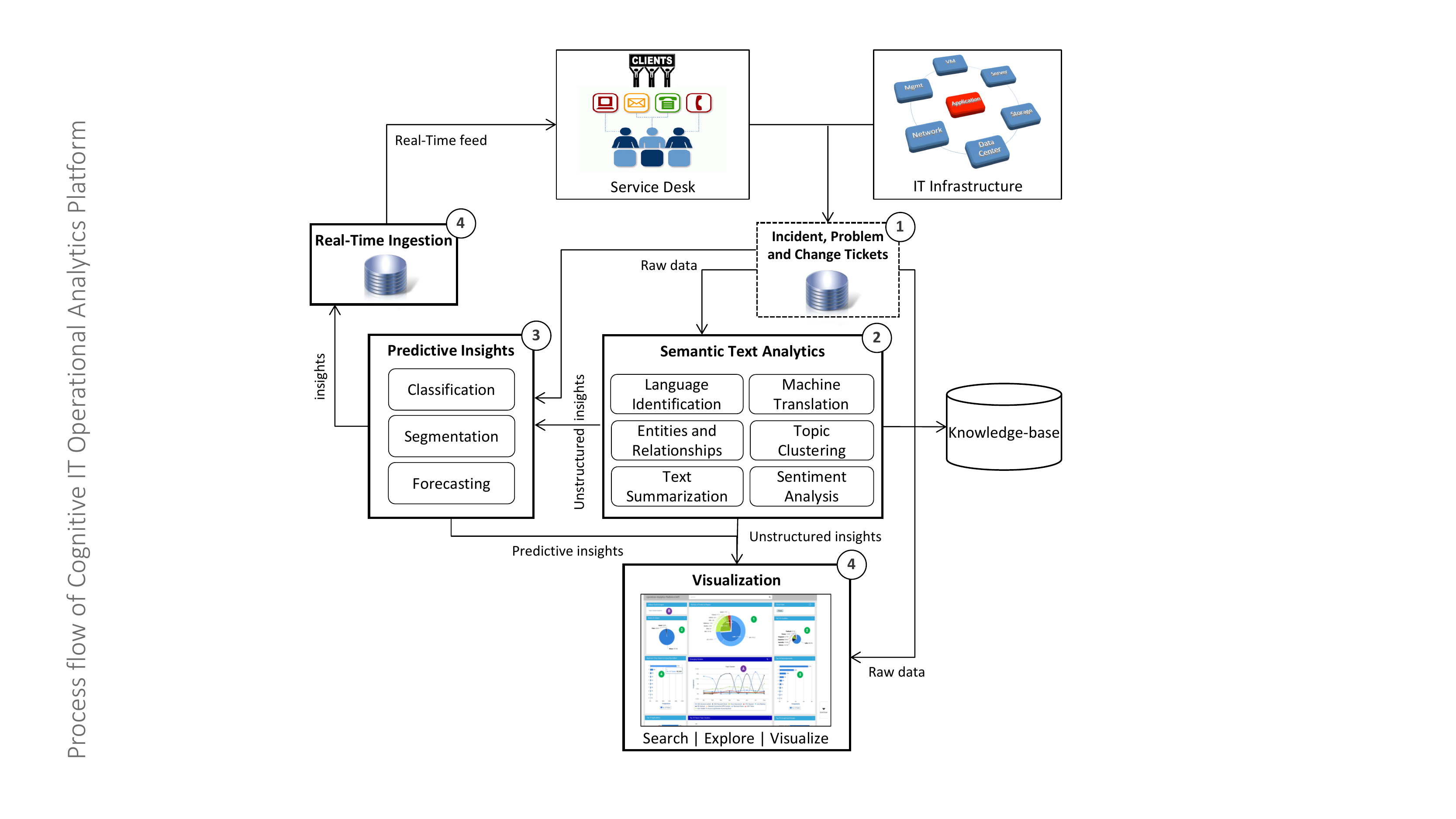}
\caption{Schematic view of Cognitive IT Operational Analytics Platform}
\label{fig:PFD}
\end{figure}

The system comprises of three layers to extract structured insights from the unstructured text as shown in Figure~\ref{fig:PFD}. The structured and unstructured data of SD is gathered in a common repository. The Semantic Text Analytics (STA) module of the system produces a pair of unique ticket identification and unstructured text attributes as input from the repository to transform into meaningful structured categories. These categories include language identification, machine translation, entities and their relationships, topic clusters, text summarization and sentiments. All these categorical features provide orthogonal dimensions of a ticket. These new features combined with the existing structured fields become an input to the Predictive Insights (PI) module. 

The PI module applies various classification, segmentation and forecasting models. The predictions, clusters and forecasting projections, combined with the raw data and text analytics insights are used to feedback the IT infrastructure and SD environments to optimize them. The visualization module is positioned to analyze a large number of tickets with their text and predictive insights. Further, the module is utilized to explore opportunities that leads to the optimization of end-to-end SD operations. This analysis cycle is adaptive where the feedback of insights is used to enhance the quality of the insights for new data which gradually improves the accuracy of the system.

\begin{figure}[!htbp]
\centering
\includegraphics[viewport=280 131 820 435, clip, scale = 0.48]{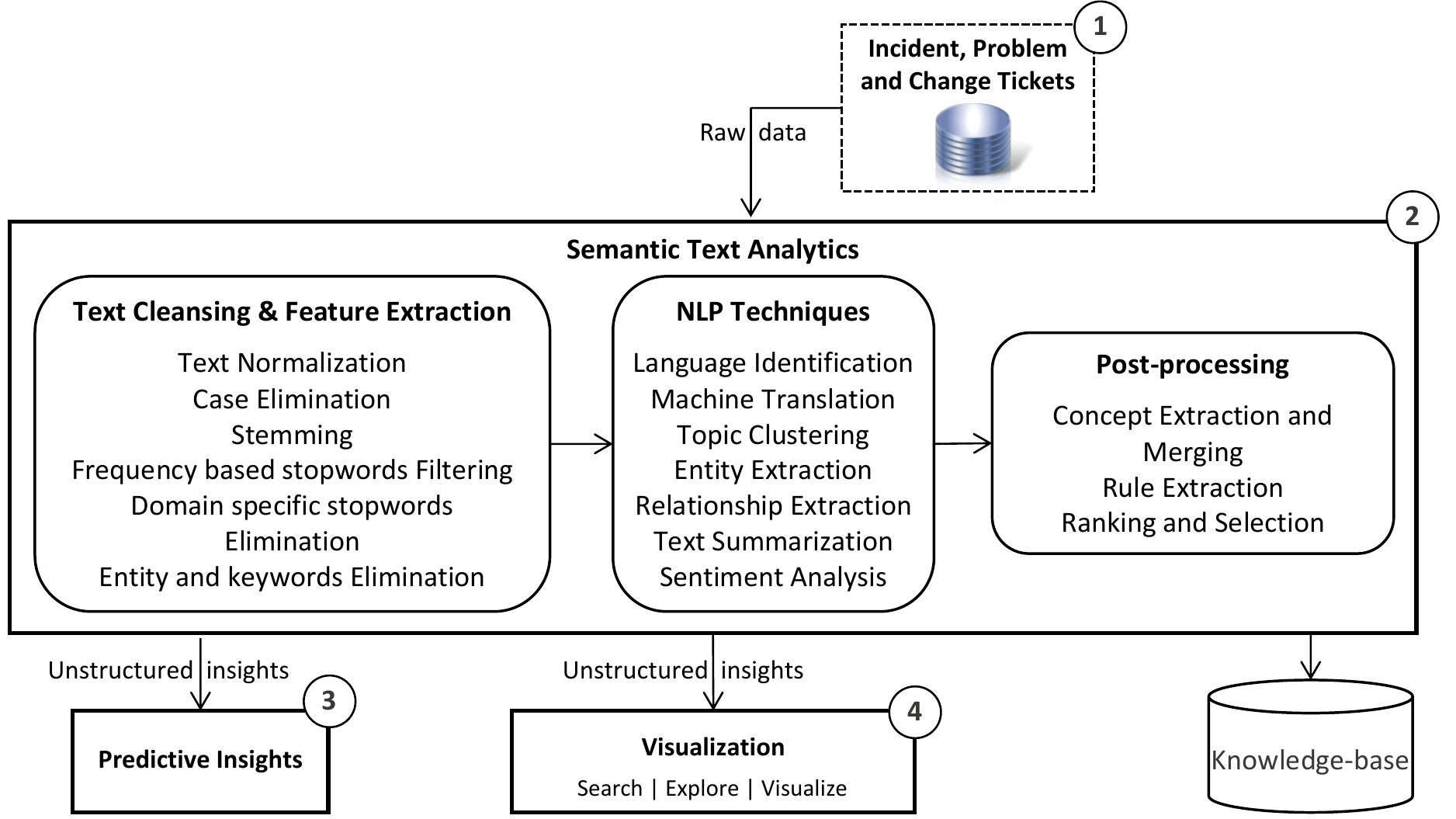}
\caption{Overview of Semantic Text Analytics}
\label{fig:STA}
\end{figure}

The STA module further comprises of three sequential layers to extract structured insights from unstructured textual data. These layers include Text Cleansing and Feature Extraction (TCFE), NLP techniques, and Post-processing layers. The high-level architecture of the module is illustrated in Figure~\ref{fig:STA}. The unstructured input text is first cleansed and normalized by eliminating its case, applying stemming, frequency and domain-specific stopwords elimination, entity elimination, and regular expression-based keywords candidate selection. The cleansed text becomes the input of NLP algorithms, however, not all the TCFE techniques need to apply as pre-processing of an algorithm. The language identification checks whether the text is in English~\cite{Jan2015}, in the case of non-English text it is translated into English~\cite{Philipp2007}. The text is further processed for hierarchical extraction of most frequent topic failures and resolutions~\cite{Osinski2004, Hofmann1999}, dynamic extraction of entities (applications, servers, printers, IP addresses, etc.) that are failing the most frequently and their relationships~\cite{Manning2014}, text summarization~\cite{Jishma2016} and sentiment analysis~\cite{Manning2014}. The STA outputs several categorical features that represent key information of the unstructured text.

The text features extracted by STA are combined with structured features of the raw data. The combined dataset has been used to predict, segment and forecast several predictive use-cases, and this module is known as PI. It consists of various use-cases which include predicting whether a newly generated ticket will violate the Service Level Agreement (SLA), identifying the most suitable (agent) skill to service the ticket, predicting Mean Time Between Failures (MTBF), propagating prevention of IT devices (servers, routers, switches, etc.) and correlating with corresponding incident ticket data. Apart from the predictive use-cases PI includes relatively simple analysis such as mean and standard deviation of hourly and peak CPU utilization of the devices. These insights feedback to SD environment to reduce the overall ticket volume and the feedback model helps to adapt the overall system.

The resulting output of STA and PI combines with raw data for interactive visualization and exploration dashboard known as Interactive Exploration and Visualization Dash-boarding (IEVD) module. Also, IEVD provides statistics and trends on the various structured data attributes of the raw ticket to identify new opportunities. The IEVD has the capability of processing massively large amounts of data and the characteristic to dynamically refresh visualization upon selection of one or several categories in real-time mode~\cite{Lahmadi2015}.

The process of taking actions to remedy the key failures and continually refreshing the findings by feeding new data is a proactive approach to maintaining a healthy IT infrastructure, increasing customer satisfaction and reducing SD operational costs. An example of such a workaround is when a network connection from a user's PC to an application server gets hung; an SD agent may not be able to determine what could have gone wrong and may suggest to the user to kill the session, reboot the machine and start a new connection with the server. While this might work for a while, further investigation at the network and server sides is suggested to identify the root cause of the problem.

\section{Algorithms} \label{sec:Algorithms}
This section uncovers the details of algorithms developed for STA, PI and IEVD. These modules are connected with each other where every module consumes the output of the ones processed earlier.

\subsection{Semantic Text Analytics} \label{sec:SemanticTA}
The TCFE component of STA has been used to remove noise and insignificant dimensions from the text. It includes space and punctuation based text tokenization, punctuation removal, case elimination, text normalization, frequency and domain-specific stopwords elimination, and regular expression-based keywords candidate selection. Furthermore, sentiment analysis based sentence filtering and shallow semantic parsing (POS tagger) based filtering have also experimented on a few datasets. These two techniques have a significant impact on the quality of the topic clustering technique where in most cases the addition of these techniques is failed to boost the quality of the clusters.

\subsubsection{Language Identification} \label{sec:LI}
There are two approaches to language identification tried, a) encoding-based and b) statistical language identification. Languages are identified by their alphabets in an encoding-based approach. Their encoding can be used to identify the languages, for an instance, in UTF-8 there is an encoding region for Japanese, Korean and Chinese languages. However, it is a challenge to distinguish European languages because their alphabet distribution overlaps with each other. On the contrary, the statistical model-based approach uses a classification framework to identify the language of the text. The two levels of modeling are required to address those languages where space is not used as word separator which includes Chinese, Korean, and Japanese. The first level is used to identify where the given text is a European or Chinese, Japanese, or Korean language where alphabet features are used for modeling framework. Accordingly, the second level further classifies that the given text belongs to which European language where word features are used in the modeling framework. Table~\ref{table:LanguageID} shows statistics around both levels of language modeling.

\begin{table}[!htbp]
\centering
\caption{Two levels statistical language modeling}
\label{table:LanguageID}
\begin{tabular}{|l|r|r|r|r|} \hline \hline
\multicolumn{1}{|c|}{\textbf{Input}} & \multicolumn{2}{|c|}{\textbf{First-level decoding}} & \multicolumn{2}{|c|}{\textbf{Second level decoding}} \\ \hline
\textbf{Language} & \textbf{Documents} & \textbf{Errors} & \textbf{Documents} & \textbf{Error} \\ \hline \hline
Spanish    &    98    &    0    &    98    &    0    \\ \hline
Danish    &    98    &    0    &    98    &    1    \\ \hline
French    &    98    &    0    &    98    &    0    \\ \hline
German    &    98    &    0    &    98    &    0    \\ \hline
Italian    &    98    &    0    &    98    &    0    \\ \hline
Portuguese    &    98    &    0    &    98    &    0    \\ \hline
Swedish    &    98    &    0    &    98    &    0    \\ \hline
Japanese    &    98    &    0    &    98    &    0    \\ \hline
Korean    &    98    &    2    &    98    &    0    \\ \hline
Chinese    &    98    &    3    &    98    &    0    \\ \hline
\end{tabular}
\end{table}

\subsubsection{Statistical Machine Translation} \label{sec:MT}
A statistical approach to Statistical Machine Translation (SMT) has been implemented by using Bayes' rule. The equation \ref{eq:1} finds most appropriate translation of a sentence in foreign language F into target language English E.
\begin{equation} \label{eq:1}
  argmax_E[P(E \mid F)] = \frac{P(F \mid E) \, P(E)}{P(F)} 
\end{equation}
The translation model calculates the probabilities of matching the source segments to the target segment by a bilingual corpus. The language model calculates the best sequences from target segments and combines them as a final output. The system has been divided into three key components a) language modeling, b) language decoding, and c) model tuning as shown in Figure~\ref{fig:STA}. The system has the capability to translate 10 non-English languages into English which are illustrated in Table~\ref{table:STAT} where the coverage and weighted coverage percentages represent the occurrence and frequency of the translated words respectively. 

\begin{figure}[!htbp]
\centering
\includegraphics[viewport=310 57 825 423, clip, scale = 0.53]{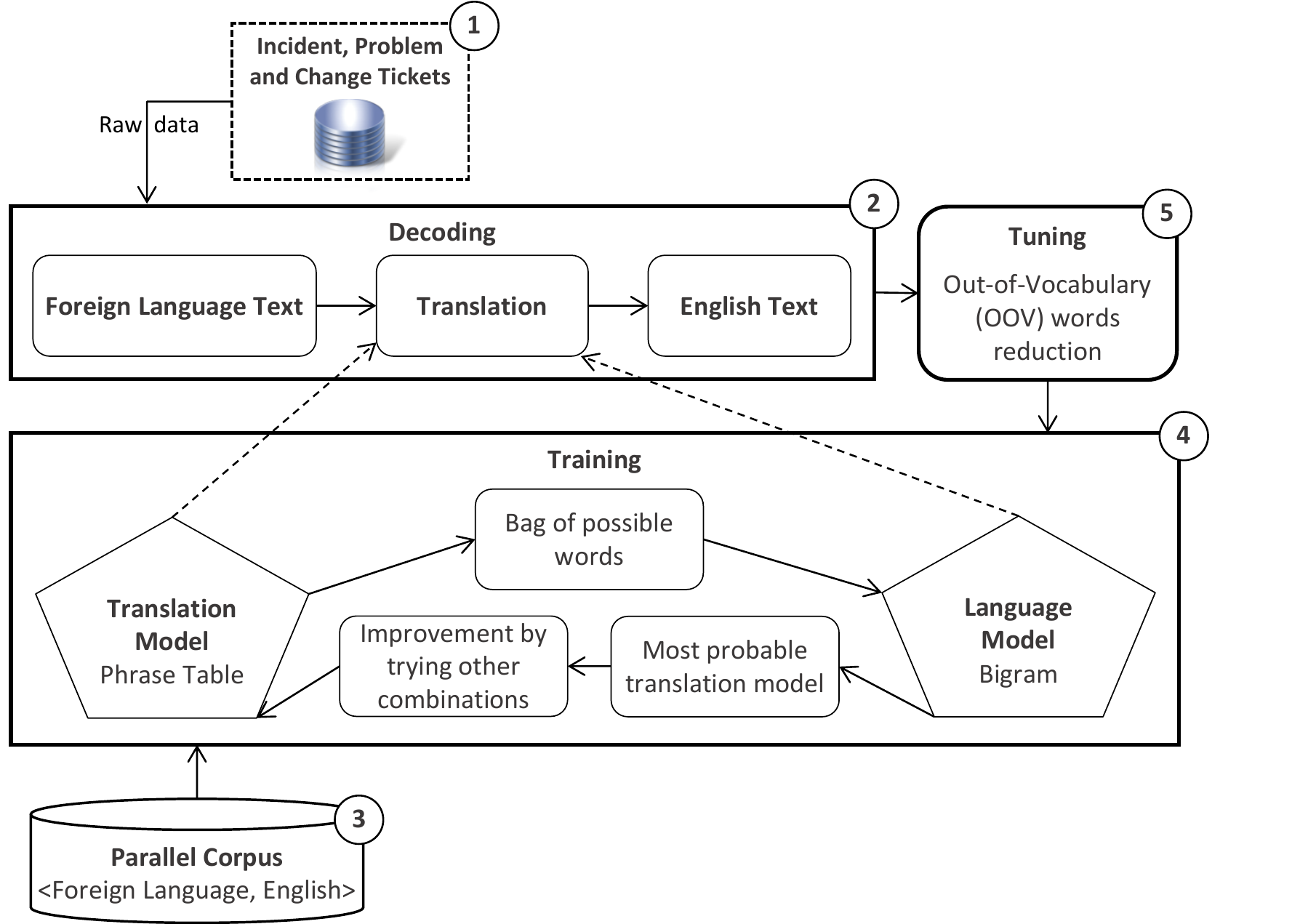}
\caption{Components of Statistical Machine Translation}
\label{fig:SMT}
\end{figure}

The system has been evaluated by first tokenizing translated English text into uni- and bi-grams with their frequency of occurrence. The most frequently occurring N-grams are filtered out and translated using another 'authentic system' with translation accuracy of 97\%+ to generate reference translation. The words which are not translated by SMT, known as out-of-vocabulary (OOV) words, are expected to be translated in reference translation. The source SMT and reference translations have been compared to compute (frequency-based) coverage and weighted coverage. Also, SMT's OOVs are translated using the same 'authentic system' and used to tune the overall translation model. 

\begin{table}[!htbp]
\centering
\caption{Statistical Machine Translation coverage and weighted coverage}
\label{table:STAT}
\begin{tabular}{|l|r|r|r|r|} \hline \hline
\multicolumn{1}{|c|}{\textbf{Input}} & \multicolumn{2}{|c|}{\textbf{Description (\%)}} & \multicolumn{2}{|c|}{\textbf{Resolution (\%)}} \\ \hline
\textbf{Language} & \textbf{Coverage} & \textbf{Wt. Coverage} & \textbf{Coverage} & \textbf{Wt. Coverage} \\ \hline \hline
Spanish    &    87.86 &    92.79 &    90.20 &    93.11    \\ \hline
Danish    &    93.22 &    92.74 & 87.34 & 89.56    \\ \hline
French    &    86.34 & 84.30 & 85.09 & 86.65    \\ \hline
German    &    90.82 & 92.45 & 89.72 & 90.44    \\ \hline
Italian    &    81.04 & 81.78 & 83.27 & 85.01    \\ \hline
Portuguese    &    86.00 &    91.01 &    88.88 &    91.35    \\ \hline
Swedish    &    84.51 & 86.03 & 85.89 & 86.71     \\ \hline
Japanese    &    75.32 & 76.10 & 79.45 & 78.45    \\ \hline
Korean    &    75.89 & 76.78 & 79.32 & 79.31    \\ \hline
Chinese    &    78.54 & 78.89 & 81.32 & 83.12    \\ \hline
\end{tabular}
\end{table}

\subsubsection{Entities and Relationships} \label{sec:ER}
Entity and Keyword extraction entail extracting tokens having any patterns. The information extracted from rules tagged under keywords consists of all caps words (application names), two or more underscores (application or server names), two or more dots (IP address), etc., whereas lexical look-ups are also made to extract domain-specific information. Consequently, the Named Entity Recognition (NER) annotator has been applied to the text to extract the information which is not picked by the rule-base and lexical look-up techniques~\cite{Jenny2009}.

The dependency relations between entities are extracted, using~\cite{Mihai2011}, to analyse the actual cause of the ticket known as relationship. It gives key summary and action elements against a ticket and filters the irrelevant details. Mostly specific elements of the ticket describe the problem or the actions taken. 

\subsubsection{Topic Clustering} \label{sec:TM}
The topic Clustering algorithm is used to uncover the cause of the problem from unstructured text data. It is the most important technique of the STA that extracts key topics of the problem description and resolution. Several algorithms have tried to come-up with an accurate topic clustering algorithm including Vector Space Model (VSM)~\cite{Peter1998}, Latent Semantic Indexing (LSI) \cite{Hofmann1999} and hierarchical N-gram clustering \cite{Osinski2004}. From different trails, VSM is found to be a less accurate technique compared to LSI and N-gram clustering. The reason is that the VSM is effected by the word usage diversity, which sometimes degrades the performance severely~\cite{Jan2015}. On the other hand, LSI works well for auto-generated logs, formulated in Equation~\ref{eq:2}, and N-gram clustering for manually generated tickets.

\begin{equation} \label{eq:2}
  P(w_j \mid D_i) = \sum_Tk {P(w_j \mid T_k) \, P(T_k)}{P(D_i \mid T_k)} 
\end{equation}
In Equation~\ref{eq:2} the words $w$ of a document $D$ consist of a set of $K$ shared latent topics $T_{1},...,T_{k}$ associated with the document-specific weights. Each topic $T_k$ in-turn offers a bi-gram distribution for observing an arbitrary word of the language. Top five most pre-dominant issues (topics), extracted from unstructured problem descriptions of various datasets, are listed in Table~\ref{table:TopicClusterLabels}.

\begin{table}[!htbp]
\centering
\caption{Most pre-dominant issues extracted via Topic Clustering}
\label{table:TopicClusterLabels}
\begin{tabular}{|l|l|l|} \hline \hline
\textbf{Dataset-1} & \textbf{Dataset-2} & \textbf{Dataset-3} \\ \hline
Low server disk space    &    Unreachable network &    Account locked    \\ \hline
Backup Scheduler    &    Related to change & VPN issues    \\ \hline
Services not running    &    Due to maintenance & Citrix session reset    \\ \hline
Faulty printer    &    Job failed  & Outlook issues    \\ \hline
Password reset    &    High CPU utilization & Access issues    \\ \hline
\end{tabular}
\end{table}

\subsubsection{Text Summarization} \label{sec:TS}
The unstructured text descriptions are typically very noisy and it is difficult to identify the part of a text containing useful information. An ontology-based approach has been implemented containing information, concepts and actions that are relevant for classification. A Term-Document-Matrix (TDM) has been generated on historical data that selects top noun and verb terms with an aggregated frequency greater than 80\%. The ontology repository has been generated with selected terms that group concepts by category including action, condition, entity, incident, negation, quantity and sentiment. These groups are annotated with weight factors, POS type and phrase type information.

\begin{figure}[!htbp]
\centering
\includegraphics[viewport=224 180 700 440, clip, scale = 0.60]{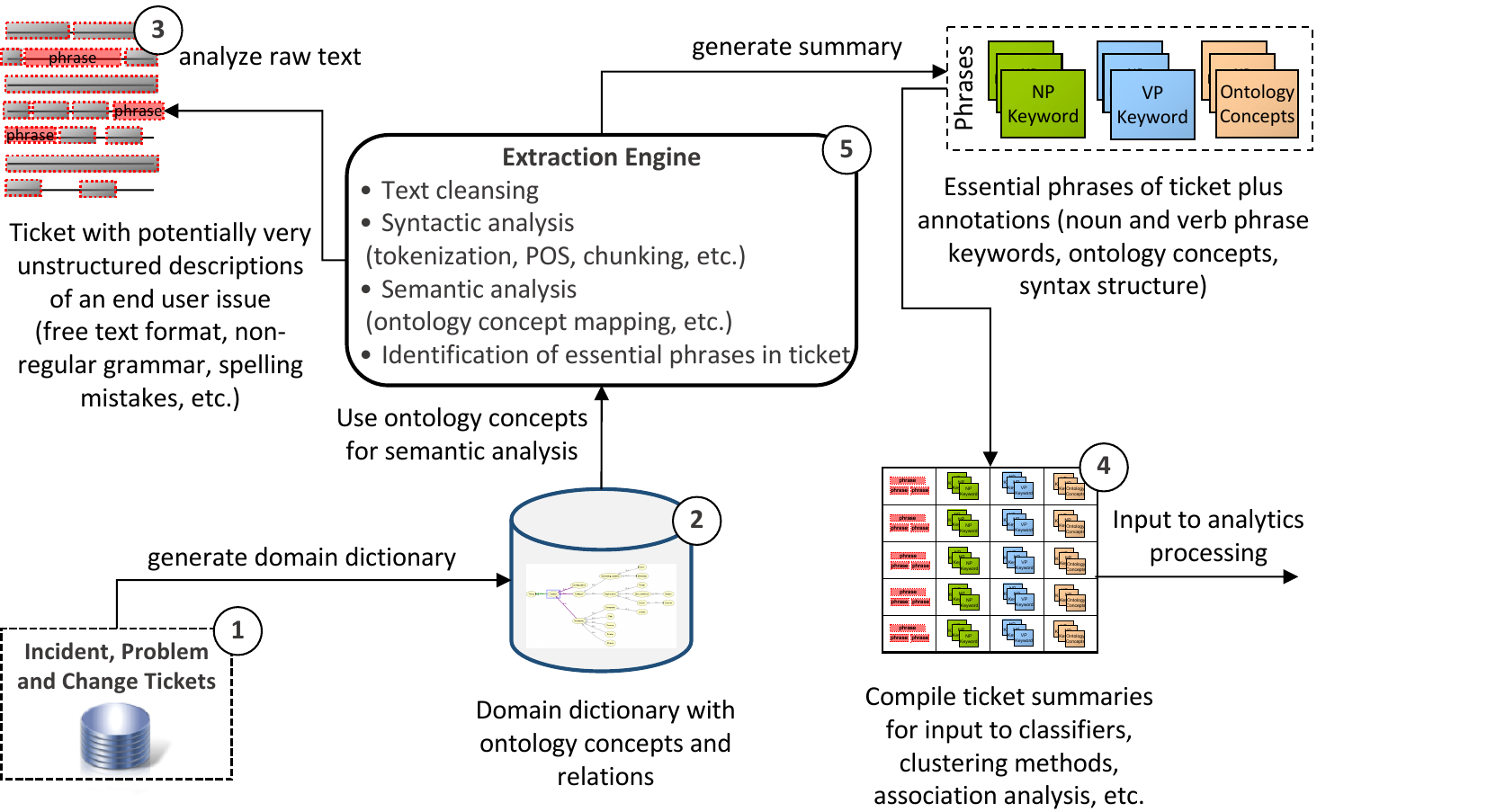}
\caption{Extraction of essential phrases using semantic mapping}
\label{fig:TextSummary}
\end{figure}

The newly generated ticket tag words with ontology concepts, similar to NER tagging and matching the POS phrase tags with corresponding ontology concept annotations. The N-gram range of 2, 5, 10, or 15 words applied on moving window over ticket description, summarize a number of matching ontology concepts and their weights for each range. Finally, the sentences with the highest density have been extracted to provide a summary that consists of summary sentences, matching ontology concepts, Noun Phrases (NP) keywords and NP + Verb Phrases (VP) keywords. Figure~\ref{fig:TextSummary} illustrates the process of extracting essential phrases.

\subsubsection{Sentiment Analysis} \label{sec:SA}
A Recursive Neural Tensor Network (RNTN) based sentiment model is used to identify the sentiments of the whole sentence instead of words~\cite{Richard2013}. However, the word-based sentiment analysis is usually carried out by defining a sentiment dictionary which then builds a tree to compute the overall sentiment. In the SD environment sentiment analysis is usually applied on CSat surreys and Social Media snippets.

\begin{figure}[!htbp]
\centering
\includegraphics[viewport=215 70 800 477, clip, scale = 0.44]{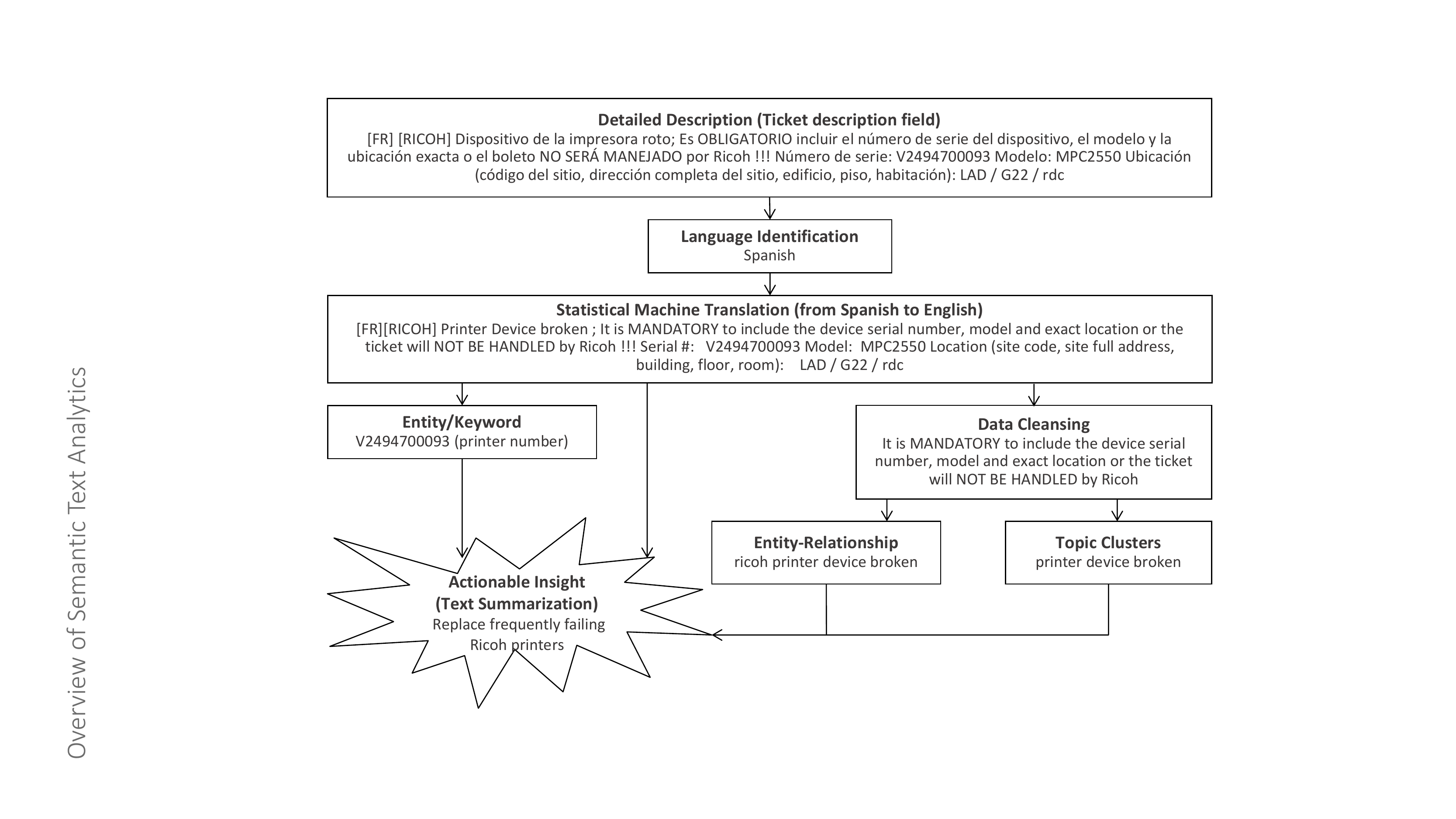}
\caption{An example how STA extracts useful information from unstructured data}
\label{fig:overview}
\end{figure}

Figure~\ref{fig:overview} shows orthogonal dimensions that STA computes from unstructured textual data. These dimensions are combined with their frequencies to find useful insights. 

\subsection{Predictive Insights} \label{sec:PI}
There are several predictive use-cases applied to the raw structured attributes and categorical insights produced by STA to classify, extrapolate and segment important features. These predictive use-cases are explained in the following sub-sections.

\subsubsection{Predict Service Level Agreement Violation}
The SLA violation is considered as a critical predictive use-case for an SD \cite{Agarwal2012}. The service level is violated when the turnaround time of a ticket exceeds the agreed duration. This delay may severely impact customer business, degrade customer satisfaction, and sometimes the vendor has to compensate the customer by paying a huge penalty in case of breaching the agreed SLA. The SLA violations can be reduced by identifying high-risk tickets which can take more time to solve than expected. Table~\ref{table:PIAccuracy} shows accuracy numbers on different datasets where the model is using both structured fields of raw data and unstructured text analytics insights.

\begin{table}[!htbp]
\centering
\caption{Accuracy of various predictive use-cases}
\label{table:PIAccuracy}
\begin{tabular}{|l|l|r|r|r|} \hline \hline
\multicolumn{3}{|c|}{\textbf{Input}} & \multicolumn{2}{|c|}{\textbf{Use-cases (\%)}} \\ \hline
\textbf{Dataset} & \textbf{Type} & \textbf{Size} & \textbf{SLA} & \textbf{OAA} \\ \hline \hline
Dataset-1    &    Incident & 267,493 & 86.41 &    79.32    \\ \hline
Dataset-2    &    Incident & 808,936 & 83.06 &    81.08    \\ \hline
Dataset-3    &    Incident & 447,740 & 86.00 &    77.73    \\ \hline
Dataset-4    &    Incident & 190,455 & 81.92 &    77.01    \\ \hline
Dataset-5    &    Incident & 65,892 & 79.03 &    75.56    \\ \hline

\end{tabular}
\end{table}

\subsubsection{Optimal Agent Assignment (OAA)}
This use-case determines the optimal agent-incident pairing to minimize the meantime to resolution (MTTR). The optimal pairing is used to smartly route incidents to the agents rather than using the traditional approach of 'first come first serve'. It reduces the overall resolution time of an incident which is explained in \cite{Ali2011}. An overall average accuracy of 83\% is recorded on four different datasets.

\subsubsection{Mean Time Between Failures (MTBF) Prediction}
The MTBF prediction use-case predicts the average time to failure of identical category tickets which fall under identical application and operating environments. It is the sequence of time of operation until the next failure occurs, which impacts both reliability and availability. Reliability is defined as the ability of SD to perform its required functions for a specified period of time. On the other hand, availability is the degree to which SD is operational and accessible.

\subsubsection{Propagation Prevention}
Propagation Prevention assesses the incidents that are likely to propagate future incidents and triggers that can be addressed to prevent such a propagation. The problem description and resolution topic clusters form associative rules which are used to determine the propagation trend among incidents. The newly arrived incidents have been flagged for the possibility of subsequent incidents where the resolution groups have been assigned to these incidents which can be proactively used to prevent propagation.

\subsection{Interactive Exploration and Visualization Dash-boarding} \label{sec:IEVD}
IEVD provides a 360-degree view of service desk tickets that harnesses all of the relevant pieces of information. It is an interactive visualization and search dashboard that can simultaneously and securely search the contents of multiple and heterogeneous datasets. IEVD also includes sophisticated mechanisms for normalization, crawling, indexing, and subsequently exploring the data \cite{Lahmadi2015}. 

IEVD is composed of three layers including Engine, Model and Presentation. The data and corresponding insights from SD repositories are crawled to the Engine layer. The Model layer creates relationships between different datasets to provide 360 degree view of SD tickets. Finally, the presentation layer provides an interactive exploration and visualization interface that allows to visually prepare query and submit it to the Engine. The data that is once crawled and indexed can be queried in real-time regardless of the volume and number of dimensions of the data.


\section{Applications} \label{sec:Applications}
This system has been deployed to optimize the service desk operations of several industries like banking, chemical, insurance, communication, retail, automotive and others. In most cases, hundreds of thousands of tickets of type incident, problem, change and SR were analyzed. Additionally, in a lot of cases, CSat data was correlated to incidents and analyzed while social media data was used for a couple of engagements. In all cases, the value ad of unstructured data analytics was demonstrated via a straight forward comparison to that of structured data analytics. It was shown how the most frequent patterns in unstructured data revealed by the system could not be possibly classified in the structured attributes of the ticket records.

In one of the engagement and for a predominant failure topic, unstructured data analytics reported a much larger proportion of tickets in that category than that of structured data analytics. In another one, statistical trends and correlation analysis revealed insights for a predominant failure topic. In the last example, a quite small proportion of incident tickets and yet quite important was reported by external clients of an organization. These incidents were backed by comments made by the external customers on social media blogs and websites. These findings got the attention of the organization who took action to resolve these issues.
The first and the foremost important and classical use-case for all accounts is the ticket volume reduction. It is pretty straightforward to extract the most predominant failure topics and sub-failure topics and their ticket counts. The recommendations are based on resolution text and IT experience, such as process automation, increasing email user notification and encouraging the use of self-service are made to the clients in order to reduce the volume of tickets.
The second use-case in importance is the correlation of incidents that propagated due to erroneous changes. In some cases, this was due to the lack of a robust test plan prior to releasing the change into a production system. In other cases, the servers on which the changes were to be made were brought down during peak work hours thereby causing users not to be able to do their work on those servers and incident creation then followed.

The other use-cases include SLA target violation and MTTR analysis. As found in the system, extensive MTTRs are often due to tickets not being transferred to the proper resolver group or not to the proper agent within the resolver group.

\section{Results} \label{sec:Results}
There are different experiments prepared to rigorously test different techniques of the STA on real datasets. These techniques are first independently evaluated and then as a single integrated system whose objective is to reduce the total ticket volume. For an instance, the techniques that are used for text cleansing and feature extraction have a dependency on the accuracy of topic clusters.

The language identification technique yields 100\% accuracy for 8 out of 11 languages mentioned in Table~\ref{table:LanguageID}. The errors in Korean and Chinese languages are due to overlapping of alphabets. However, the system is more biased towards European languages. If more than 20\% of Asian language input contains English, the system decodes it as the European language.

\begin{figure}[!htbp]
\centering
\includegraphics[viewport=290 45 670 465, clip, scale = 0.67]{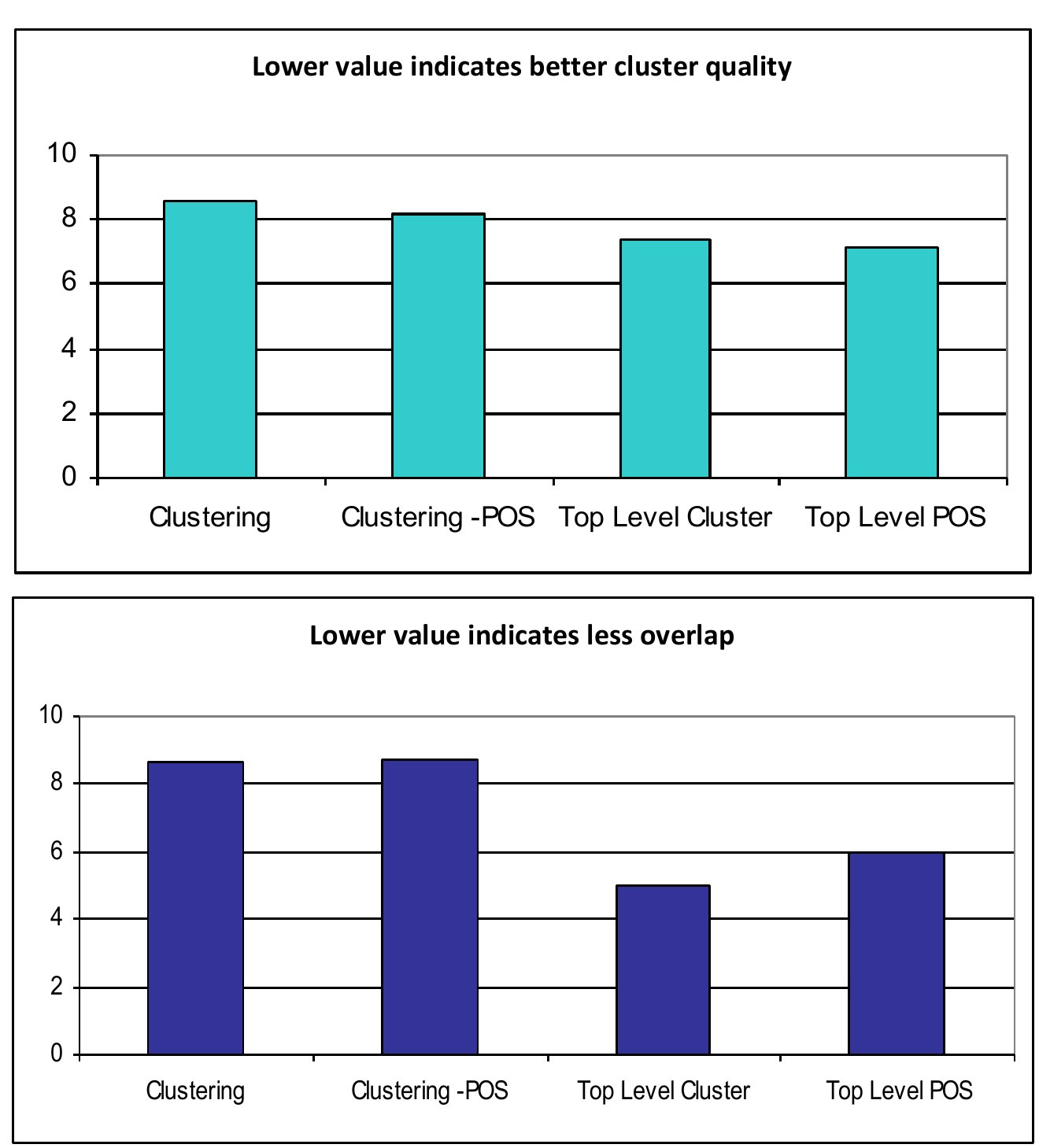}
\caption{POS Based Ticket Feature Extraction}
\label{fig:TopicClusteringAnalysis}
\end{figure}

The machine translation of 10 different languages into English is listed in Table~\ref{table:STAT}. The non-English SD problem description and resolution text have been translated into English where the coverage and weighted coverage measures are used to evaluate the system. The results show consistently better performance on European and Latin-American languages as compared to the Asian languages.

\begin{table}[!htbp]
\centering
\caption{Topic Clustering coverage and weighted coverage}
\label{table:TM}
\begin{tabular}{|l|l|r|r|r|} \hline \hline
\multicolumn{3}{|c|}{\textbf{Input}} & \multicolumn{1}{|c|}{\textbf{Description (\%)}} & \multicolumn{1}{|c|}{\textbf{Resolution (\%)}} \\ \hline
\textbf{Dataset} & \textbf{Type} & \textbf{Size} & \textbf{Coverage} & \textbf{Coverage} \\ \hline \hline
Dataset-1    &    Incident &    267,493 &    67.36 &    74.60    \\ \hline
Dataset-2    &    SR &    173,368 &    70.64 &    77.78    \\ \hline
Dataset-3    &    Incident &    808,936 &    92.84 &    61.97    \\ \hline
Dataset-4    &    Incident &    447,740 &    72.86 &    75.68    \\ \hline
Dataset-5    &    Incident &    65,892 &    63.49 &    81.90    \\ \hline

\end{tabular}
\end{table}

Topic clustering is found to be one of the most challenging and critical techniques of STA. There are several experiments designed to test the best topic clustering technique that can persistently give the best results on a different dataset. Apart from the selection of appropriate algorithm, different pre-processing techniques have been tested as it is highly dependent on how the input data has been cleansed. The two key techniques which are used to clean the data include sentiment analysis and part-of-speech (PoS) phrases. The idea is to extract negative and noun-verb phrases from noisy data using sentiment analysis and PoS respectively. The sentiment analysis based sentence filtering cleansing technique reduced the overall clustering accuracy and coverage. However, PoS phrases extraction, as a pre-processing technique, helps in boosting topic clustering quality in most of the cases where it provides a key phrase describing problem symptoms and actions taken. As the phrases are getting small, clustering results are showing marginal improvement. The results of the cluster evaluation are shown in Figure~\ref{fig:TopicClusteringAnalysis}. The overall topic clustering results are summarized in Table~\ref{table:TM} for both problem description and resolution.

\begin{figure}[!htbp]
\centering
\includegraphics[viewport=242 60 850 450, clip, scale = 0.45]{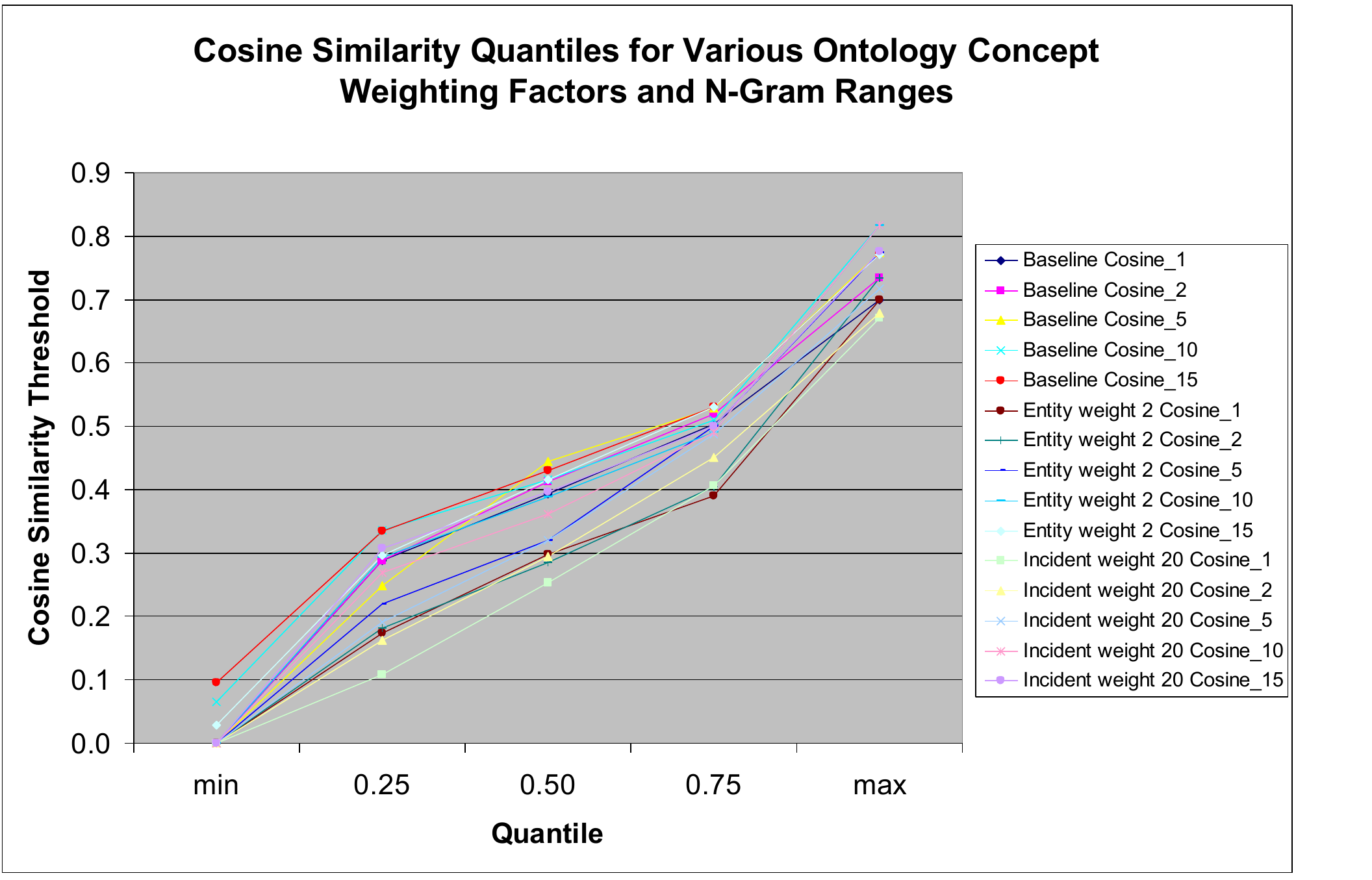}
\caption{Cosine similarity quantiles for various ontology concepts of Text Summerization}
\label{fig:TextSummaryAnalysis}
\end{figure}

Text Summarization becomes very useful for both summary sentences and NP-VP phrases. In a few trails, manually labeled test-set have been used to calculate cosine similarity for various ontology concept weighing factors and N-gram ranges using 'moving window size'. The N-gram range of 15 words has the highest similarity to manually labeled sentences in the test-set which can be visualize in Figure~\ref{fig:TextSummaryAnalysis}.

\section{Conclusion} \label{sec:Conclusion}
The SD ticket dispatch system plays a vital role to provide persistent IT services. The proposed system provides a sequential multi-component semantic analysis platform that combines the insights of a number of text and predictive analytics techniques to accurately solve a complex task. This platform extracts information from both unstructured and structured data for deep analysis and interactive visualization. The three-layer methodology is found to be very effective in which step-by-step extracts complex information from a layer where insights computed by all previous layers become its input to get richer insights. The usage of various NLP techniques on a ticket provides orthogonal dimensions of problem areas which eventually addresses a few key challenges. These challenges include the extraction of ambiguous problem areas of the tickets and low coverage and accuracy of individual techniques. Many organizations are using this system and they have observed improvements not only in their SD operations but in the stability of their IT infrastructures as well. All the text and predictive analytics techniques that are part of this system are giving significantly higher and consistent accuracy on different types of cross-domain noisy datasets. There is currently on-going work to enhance this system by making its data processing even faster and incorporating new algorithms for a time-based correlation between two or more related data sets.


\end{document}